\documentclass{article}

\usepackage{PRIMEarxiv}

\usepackage[utf8]{inputenc} 
\usepackage[T1]{fontenc}    
\usepackage{hyperref}       
\usepackage{url}            
\usepackage{booktabs}       
\usepackage{amsfonts}       
\usepackage{nicefrac}       
\usepackage{microtype}      
\usepackage{lipsum}
\usepackage{fancyhdr}       
\usepackage{graphicx}       
\graphicspath{{media/}}     
\usepackage{cite}
\usepackage{amsmath,amssymb,amsfonts}
\usepackage{textcomp}
\usepackage{algorithmicx}
\usepackage{algorithm}
\usepackage[noend]{algpseudocode}
\usepackage{multirow}
\usepackage{array}
\usepackage{float}
\usepackage{caption}
\usepackage{subcaption}
\title{Equitable-FL: Federated Learning with Sparsity for Resource-Constrained Environment
}

\author{
  Indrajeet Kumar Sinha \\
  Department of Information Technology\\
  Indian Institute of Information Technology Allahabad\\
  Prayagraj, Uttar Pradesh, India. \\
  \texttt{pcl2016004@iiita.ac.in} \\
   \And
  Shekhar Verma \\
  Department of Information Technology\\
  Indian Institute of Information Technology Allahabad\\
  Prayagraj, Uttar Pradesh, India. \\
  \texttt{sverma@iiita.ac.in} \\
  \AND
  Krishna Pratap Singh \\
  Department of Information Technology\\
  Indian Institute of Information Technology Allahabad\\
  Prayagraj, Uttar Pradesh, India. \\
  \texttt{kpsingh@iiita.ac.in} \\
}

\begin{document}
\maketitle

\begin{abstract}
In Federated Learning, model training is performed across multiple computing devices, where only parameters are shared with a common central server without exchanging their data instances. This strategy assumes abundance of resources on individual clients and utilizes these resources to build a richer model as users’ models. However, when the assumption of the abundance of resources is violated, learning may not be possible as some nodes may not be able to participate in the process. In this paper, we propose a sparse form of federated learning that performs well in a Resource Constrained Environment. Our goal is to make learning possible, regardless of a node’s space, computing, or bandwidth scarcity. The method is based on the observation that model size viz a viz available resources defines resource scarcity, which entails that reduction of the number of parameters without affecting accuracy is key to model training in a resource-constrained environment. In this work, the Lottery Ticket Hypothesis approach is utilized to progressively sparsify models to encourage nodes with resource scarcity to participate in collaborative training. We validate Equitable-FL on the $MNIST$, $F-MNIST$, and $CIFAR-10$ benchmark datasets, as well as the $Brain-MRI$ data and the $PlantVillage$ datasets. Further, we examine the effect of sparsity on performance, model size compaction, and speed-up for training. Results obtained from experiments performed for training convolutional neural networks validate the efficacy of Equitable-FL in heterogeneous resource-constrained learning environment.
\end{abstract}


\vspace*{-\baselineskip}
\section{Introduction}
\label{introduction}
\begin{figure}[H]
    \centering
    \includegraphics[trim=300 135 320 110, clip, scale=0.60]{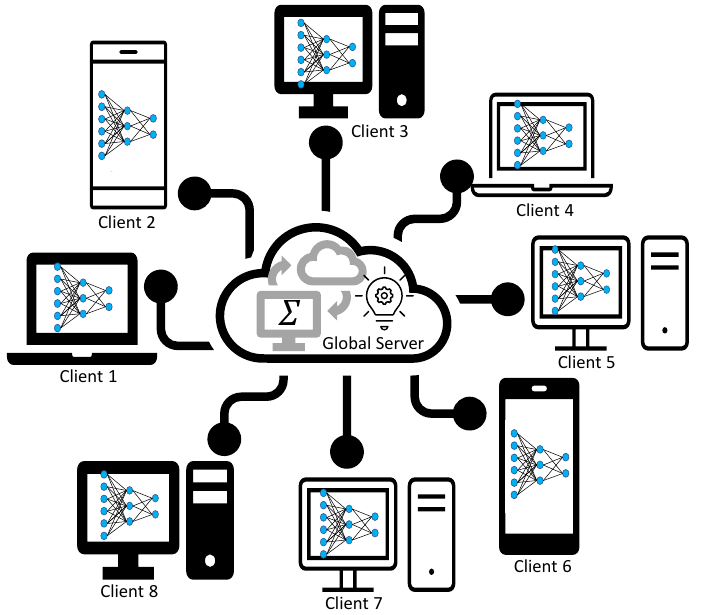}  
    \caption{\textbf{Overview of Federated Learning Architecture. The global server has a connection with all clients. Clients are equipped with personal data, computation power and training models. They can share updates to the server for computation of the global model.}}
    \label{federated_learning_overview}
\end{figure}

Centralized supervised and unsupervised machine/deep learning algorithms have achieved state-of-the-art performance in domains like speech recognition, natural language processing and computer vision (\cite{deep_survey_3}, \cite{unsupervised_learning_1}, \cite{deep_learning_1}). However, these models require localizing training data on a central repository/pool \cite{deep_learning_1}. In application domains like healthcare, military and space communication that feature sensitive and private data, this localization of sensitive data on a central repository is not feasible as data owners are uncomfortable sharing their sensitive data. In this way, even though data is available with the data owners, its utilization in training state-of-the-art deep learning models is restricted (\cite{federated_healthcare}, \cite{federated_healthcare_2} \cite{federated_learning_survey_2}). Federated learning ($FL$) overcomes this limitation by allowing the data owners to retain/not share their sensitive data and still be able to participate in collaborative training to obtain a workable global model \cite{federated_learning_survey_1}. This setup has a global server and several participating clients/data owners/organizations. The data available with the individual clients are retained, and local models are trained at the client level. The knowledge obtained at the client level is then communicated to the global server that, in some sense \textit{aggregates} this knowledge and sends it back to the participating clients. The participating clients then resume their training starting with these \textit{aggregated} knowledge. This communication of local knowledge from participating clients to the global server and the subsequent transfer of \textit{aggregated} knowledge from the global server to the participating clients constitutes one communication round. The overview of this setup is highlighted in Figure \ref{federated_learning_overview}. In this way, the otherwise non-usable data (due to the reluctance of data owners to share their data with the central pool) participates in model training without compromising security and privacy (\cite{federated_learning_survey_2}, \cite{federated_survey4}). \\

Several approaches are proposed to train deep learning models in the federated setup. Specifically, these approaches fall in one of the three settings - vertical setup, horizontal setup and hybrid setup depending on what information is shared among the clients \cite{federated_taxonomy}. In the horizontal setup, the feature space is shared among all the clients and the sample space is divided. Vertical setup divides the feature space among the clients and retains the sample space; hybrid setup is a mix of vertical and horizontal setup and allows sharing of feature and sample space among the participating clients. Each of these settings has its challenges. In the horizontal setup, the clients with large samples may start to dominate the collaborative training. In the vertical setup, the main challenge is to unify the information in the disjoint feature spaces and perform domain adaptation at the client level. A hybrid setup may have both these limitations. In addition, little attention is given to regularizing the network (in terms of total trainable parameters) to accelerate federated learning (\cite{federated_taxonomy}, \cite{federated_learning_survey_2}). This reduction in the network size is important, as the participating clients have limited energy resources available for training the local model. Moreover, the network's reduced size, parameter space, latency, and associated communication cost can reduce \cite{federated_taxonomy}. \\

Most federated learning-based approaches use dense models to train state-of-the-art models; these approaches assume that all the neurons in the network are significant in learning feature representations and should fire. However, with the reduced sample spaces, particularly in the horizontal and hybrid setting, this assumption may not hold \cite{federated_taxonomy}. Consequently, dense networks in these settings are prone to overfit the training data \cite{DeepL}. The parameter space remains high dimensional and adversely affects the total communication cost across multiple rounds. Sparse models in the federated-learning setup can overcome these limitations by learning the generalized feature representations and keeping the parameter space small \cite{sparse_deep_1}. \\

Several sparse models have been proposed in application domains such as computer vision, natural language processing, computational biology and representation learning to solve tasks like image recognition, feature representation, speech recognition, image translation and image tagging (\cite{sparse_deep_1}, \cite{sparse_deep_2}, \cite{sparse_deep_3}, \cite{Sparsity-in-Deep-Learning}). However, limited attention has been given to investigating sparsity's effect on the model's generalisation capability \cite{Sparse-Networks-from-Scratch}. Additionally, the sparse models proposed in the federated learning setup give limited attention to using sparsity for reducing communication costs across multiple rounds (\cite{federated_sparse1}, \cite{federated_learning_survey_2}, \cite{federated_survey4}).  \\

\vspace*{-\baselineskip}

This work demonstrates the possibility of training sparse networks that yield good generalization in federated architecture. We divide the work into three phases. In the first phase, we include only 30\% of the total client formation considering these clients as highly efficient. Hereupon, training of dense networks continues till the accuracy of the test set stops increasing. Once the dense network is trained, we enter into the second phase. Now, p\% of the smallest-magnitude weights are sparsed, and the remaining weights are reset with the initial weights. The Lottery ticket hypothesis is as per \cite{malach2020proving}. Considering the moderately efficient clients, we include the next 30\% of the total client along with previous ones. Afresh training starts, maintaining the sparsity level of the p-sparse network in federated architecture till validation performance stops increasing and after reaching the stopping condition for this phase. In the last phase, allowance is made for the weaker clients and all clients are incorporated. Following the Lottery ticket hypothesis, the network's sparsity is now increased. Again, training maintains the new sparsity level. Training stops after obtaining desired performance. We test our approach with five data sets demonstrating a progressive sparsifying network dealing with heterogeneous networks. The significant contributions are outlined as follows:
\begin{itemize}
    \item A novel federated learning framework for the constrained environment with resource heterogeneity that achieves higher speedup and lower communication overhead in different settings. This is achieved by training the model through multiple phases of training and sparsification. Nodes with higher heterogeneity score participate in earlier phases as compared to weaker nodes with lower heterogeneity score.
    \item Several datasets are formed to support FL under heterogeneous settings. We show a communication overhead in kilobytes that reduces with the increase in network sparsity.  
    \item A metric to evaluate speedup regarding floating-point operations to quantify the reduction of computations across clients.
    \item The model is implemented on the healthcare $Brain-MRI$ image dataset \cite{BMRI} and the $PlantVillage$ dataset \cite{PlantData}, and its efficacy is demonstrated for the resource-scarce heterogeneity FL environment.
\end{itemize}

The rest of the paper is structured as follows. Section 2 describes the Equitable-FL together with the algorithm. Section 3 discusses experimental conditions such as model architecture and datasets. Section 4 examines the performance of the suggested model. Section 5 takes the paper to a conclusion.
\section{Problem Description}
Resource heterogeneity and scarcity are major issues in federated learning, which make the learning environment inconsistent. For example, a client may have very slow network access, or another may need more computational capacity to work on a complex model. As a consequence, the participation of a node may become unreliable, or an active node may drop out anytime.
As a result of limitations indicated on each device, only a fraction of the nodes normally participates in the learning process. Leaving out such straggler nodes may lead to extremely low model quality. Existing FL approaches are only capable of handling a part of the problems of the paucity of resources and heterogeneity. Resource scarcity and heterogeneity may lead to node nonparticipation, cause large time delays in model updates or result in a substantial accuracy-model size tradeoff. Due to limitations on devices and bandwidth limitations across a potentially large network, a different approach is needed to overcome heterogeneity and resource paucity issues. It is observed that the main problem is the large model size, which entails the presence of huge computing and memory on each node for training a model with large number of parameters that need to be updated and exchanged between a node and the global server. Thus, it is necessary to reduce the number of parameters of a model to match the resource budget at a node. However, this should not degrade the model quality. 

\section{Proposed Method}

The proposed method is based on the two observations. One, the problems of scarcity and heterogeneity arise due to large model size and two, it is possible to reduce the model size without compromising model performance. We present a progressive FL framework with the sparsification strategy for dealing with heterogeneity and resource constraints. This is achieved by restricting the model complexity by reducing the number of parameters without sacrificing accuracy. The approach allows training complicated models on any device, regardless of its space, computing, or bandwidth limitations. The method enables nodes to participate in training when heterogeneity information is known a priori. The selection of nodes in the Equitable-FL is based on heterogeneity score of a node. Heterogeneity score is the minimum of resources, computation and storage, on a node and network bandwidth available to the node for communication. The heterogeneity score reflects the resource bottleneck of a node. Thus, all nodes with high heterogeneity score can participate in a few first training rounds. Concurrently, the server starts reducing the number of parameters in the global model to increase its sparsity, encouraging nodes with lower heterogeneity score to participate in training in later phases when the model becomes sparse. 
\vspace*{-\baselineskip}
\subsection{Notation}
        \vspace*{-\baselineskip}
        \begin{table}[H]
            \begin{center}
            \begin{tabular}{|c|c|c|}
                \hline
                \textbf{S.No.} & \textbf{Notations} & \textbf{Descriptions} \\ \hline
                 1 & $N$                &  Total number of clients \\ \hline
                 2 & $k$                &  Clients selected for aggregation at Server \\\hline
                 3 &  $\alpha$          &  Learning rate of the optimizer \\\hline
                 4 &  $d_{i}$           &  Local data at $i^{th}$ client \\\hline
                 5 &  $m_{i}$           &  Cardinality of local data \\\hline
                 6 &  $M$               &  Size of k-client data for aggregation\\\hline
                 7 &  $\l(.)$   &  Loss function\\\hline
                 8 & $\Theta^{*}$           & Server's final weight \\\hline
                 9 & $\Theta_{r}$           &  Server aggregated weight at $r^{th}$ round \\\hline
                 
                 10 & $\theta_{r}^{i}$         &  Local weight of $i^{th}$ client at $r^{th}$ round \\\hline 

                 11 & $p$         &  Amount of Sparcification in percentage   \\\hline 
                 12 & $prune\_rate$    &  Ratio for sparsifying weights   \\\hline 
                 13 & $R[phase]$         &  Total number of rounds for distinct phase   \\\hline 
                 14 & $flag$         &  Variable that is a condition for execution of the LTH   \\\hline 
                 15 & $clients\_pool$         &  Group of clients with a threshold of heterogeneity score  \\\hline 
                 16 & $prune\_mask$     &   Binary mask that
                 disables the gradient flow  \\\hline 
                 17 & $|\Theta|$     &   The total number of parameters in weight \\\hline
                 18 & $||\Theta ||_{0}$     &   The total number of nonzero parameters in weight  \\\hline
        
            \end{tabular}
            \end{center}
            \caption{\textbf{Description of different notations used in context to the Equitable-FL.}}\label{table:note}
        \end{table}
        \vspace*{-\baselineskip}
\subsection{Federated Learning} 
            In federated learning, multiple clients train machine learning models collaboratively without exchanging their local data \cite{pmlr-v54-mcmahan17a}. The outcome of this technique is a global model available for all clients. In a practical federated learning system, the global model performs better than the model learned by local training with the same model architecture. FL can be expressed mathematically as:
            \begin{equation}\label{eq:FL}
                \Theta^* = \underset{\Theta}{ \operatorname{argmin}} \hspace{0.6em} \sum_{i=1}^{k}\frac{m_i}{M} {\l_i}(\Theta)
            \end{equation}
            
            Here, $\l_{i}$(·) expresses the loss of global model parameters $\Theta$ on local data at $i^{th}$ client.

\subsection{Sparsification}
            Sparsification is a method to produce sparse neural networks. A vector is $k-sparse$ if its $\l_0$ norm is $k$ or it has no more than k non-zero entries. Similarly, neural networks are classified as sparse or dense when the activation of neurons within a specific layer is supervised. "Sparse connectivity" is a condition where only a small subset of units are interlinked. A link with zero weight is effectively unconnected, similar to sparse weights.
  
            We drop the $p$ connections in the neural network once throughout the training stage. This $p$ is calculated as mentioned in the equation \ref{eq:Prune}.

            \begin{equation}\label{eq:Prune}
                p = (|\Theta| -||\Theta ||_{0}) + prune\_rate * ||\Theta||_{0} \\
            \end{equation}
            \vspace*{-3em}
\subsection{Lottery Ticket Hypothesis}
            The Lottery Ticket Hypothesis is an approach for training neural networks such that a certain subnetwork is reinitialized and the remaining subnetwork is detached after its conventional training can match the original network's test accuracy after training at most the same number of iterations \cite{malach2020proving}. It focuses on sparsifying weights so that certain sparsed subnetworks can be retrained to perform similarly to the whole network. \cite{LTH2018}
            stated \textit{the lottery ticket hypothesis} as \textit{"A randomly-initialized, dense neural network contains a subnetwork that is initialized such that—when trained in isolation—it can match the test accuracy of the original network after training for at most the same number of iterations"}. When the right subnetwork is selected, accuracy is unaffected, while the network size can shrink, evolves faster, and consumes fewer resources during inference. 
            \vspace*{-\baselineskip}
\subsection{Speed-up }
            The degree of sparsity decreases the count of floating point operations required for the lighter model.
            
            Taking sparse model parameters matrix as $W$ and input vector as $x$. Computing $(Transpose(W) * x)$ requires computing only the pointwise product at the indices $W_{i,j} \neq 0$ and $x_{j} \neq 0$. Sparsity ensures that floating-point operations are minimized and reasonably speeds up primitive processors of limited parallelization and cache memory. Speed-up measures the relative performance of two approaches to the same problem. Speed-up is generally used to show performance effects after any architecture enhancement. The following formula defines speed-up in floating point operations(FLOPs):
            \begin{equation}\label{eq:Speed-up}
                     \begin{split}
                            S_{FLOPs} = \frac{FLOPs_{dense}}{FLOPs_{sparse}} \hspace{10.3em}\\ 
                                    = \frac{Total\hspace{0.15em}number\hspace{0.15em}of\hspace{0.15em}Parameters}{Total\hspace{0.15em}number\hspace{0.15em}of\hspace{0.15em}Nonzero\hspace{0.15em} Parameters} \\
                      \end{split}
            \end{equation}
\vspace*{-\baselineskip}
\subsection{Space-saving}
            The lossless compression technique reduces the size of data without removing any information. Restoring the file to its prior state without losing data during decompression is possible. We used RAR as an archive file format for lossless data compression for file sharing with fewer resources. Space-saving refers to the quantity of space saved as a result of compression. 
            \begin{equation}\label{eq:Space-saving}
                {SpaceSaving} = 1 - \frac{Size(SparseModel)}{Size(DenseModel)} \\
            \end{equation}
            
\subsection{{Equitable-FL}}
Algorithm \ref{alg:Client Federated} illustrates federated learning with sparse models. This method differs from vanilla federated architecture and it executes in three phases. It initially allows dense network learning on locals with progressive sparsification later. Algorithm \ref{alg:server Federated} depicts the server functions. The server communicates with the clients, controls learning, and determines when to apply the LTH to the parameters. Algorithm \ref{alg:prune} depicts the LTH approach used to sparsify weights at the beginning of the second and third phases on the server side.
\begin{figure*}[hbt]
    \includegraphics[trim=0 215 0 20, clip, scale=0.49]{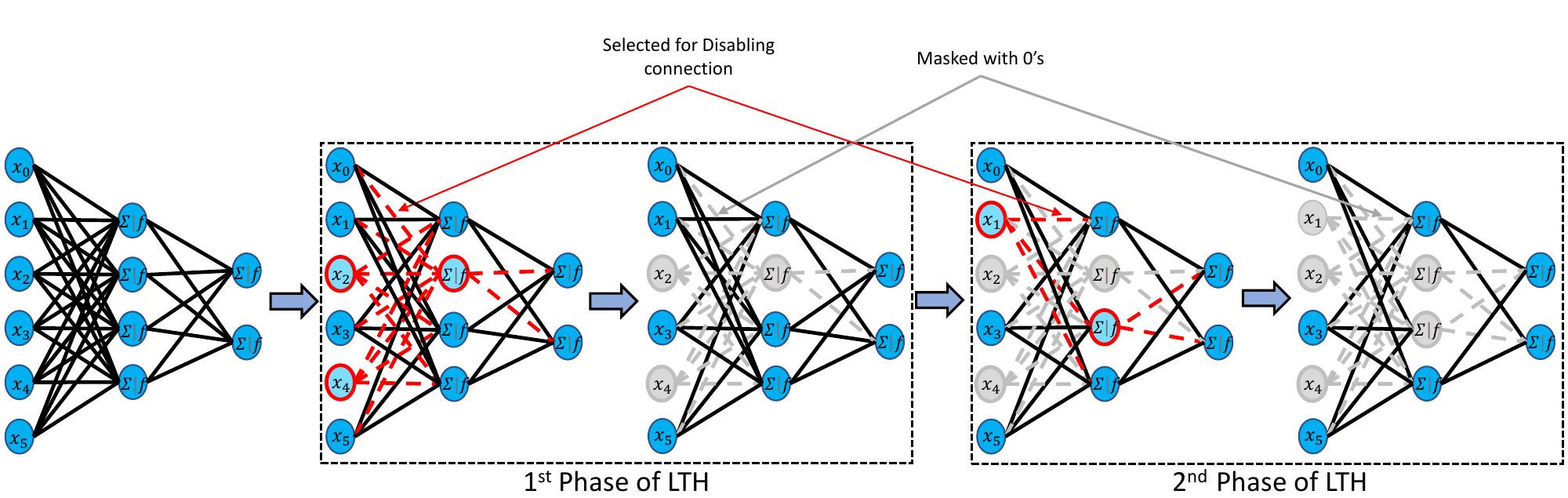}  
    \caption{\textbf{Overview of the Equitable-FL as detailed in Algorithm 1. The neural network is initially dense; as the training progresses, the connections between some neurons are dropped, also called sparsifying using the LTH approach. This process is performed in two phases to make it possible for the participation of clients with low heterogeneity score.}}
\end{figure*}
Each round of client training in a federated framework is illustrated in Algorithm \ref{alg:Client Federated}. Three groups(say clients\_pool) of clients are formed according to their resource capacity; the first group has highly capable clients only, the second group has extra clients with the first group with moderated capable, and the third or the last group has least capable clients in addition to second group clients. The entire training is executed in three phases on respective clients\_pool. All clients are initially configured with the same Server parameters ($\Theta_o$). Then, clients begin training the model using their local data ($d_i$). Following training, each client gets its trained parameters, which are then sent to the server, and the server chooses any $k$-clients from an n-client set for aggregation. We chose $k = 2$ as the value. We refer to client-side training and server-side aggregation as one-round communication. By aggregating k-selected parameters, the server computes global parameters. This training process for one phase continues till R[phase] number of rounds. When the client's training is completed for the first phase, the same procedure is again followed for the next phases. However this time, sparsity is maintained at the client's training using the server's mask.

\floatname{algorithm}{Algorithm}
\renewcommand{\algorithmicensure}{\textbf{Outcome:}}
\newcommand{\End}{\textbf{END}}
\begin{algorithm}[H]
    \caption{LTH training client-side}\label{alg:Client Federated}
    \begin{algorithmic}[1]
        \Require {$\Theta_{r}$, $\alpha $, $prune\_mask$ ; \textit{Client\hspace{2px}i\hspace{1px}}: ${d_{i}} \in {D}$}
        \Ensure Local $i^{th}$client Updates: ${\theta_{r+1}^{i}}$
        \raggedright

        \hspace{-1em}\textbf{for} $phase$ in len(clients\_pool) \textbf{do} \\
        \hspace{0em} \textit{set rounds to \textbf{R[phase]}}\\
        \hspace{0em} \textit{set r to \textbf{zero}}\\
        \hspace{0em} \textbf{while}($r$ $<$ rounds) do \\
        \hspace{1.8em} \textbf{for} $i$ in clients\_pool[phase] \textbf{do} \\
         \hspace{3em} // \textit{\textbf{local client's parameter initialization.}} \\
         \State     \hspace{3em} ${\theta_{r}^{i} \leftarrow \Theta_{r}}$ \hfill
        \raggedright 

        \hspace{3em} // \textit{\textbf{Training at $i^{th}$client on their dataset$(d_{i})$.}} \\
        \State\hspace{3em}  {\textbf{if} $phase ==$ "1" {\textbf{then}} }\\ \hspace{1.5em} 
                    \hspace{4em} ${\theta_{r+1}^{i}} \leftarrow \theta_{r}^{i} - \alpha {\nabla}\l(d_{i},\theta_{r}^{i}) $ \\ 
        \State\hspace{3em}  {\textbf{else}  }\\ \hspace{1.5em} 
                    \hspace{4em} ${\theta_{r+1}^{i}} \leftarrow \theta_{r}^{i} - \alpha {\nabla}\l(d_{i},\theta_{r}^{i}) \odot prune\_mask $ \\ 
        \raggedright
        \vspace{0.2em}
        
        \State \hspace{3em} \textit{\textbf{Client shares parameters to Server.} } 
        
        \hspace{2em} \textbf{end for} 
        
        \hspace{0.5 em} \textbf{end while}   \newline
        \hspace{-1em}\textbf{end for}
    \end{algorithmic}
\end{algorithm}

Algorithm \ref{alg:server Federated} shows the server-side operation. The server performs differently in three phases. It decides clients participation in phase-wise training and performs parameter aggregation. For the second and third phases, it applies the LTH, $p\%$ of the smallest-magnitude weights are sparsed, and the remaining weights are reset with the original initial weights and communicated to all clients. Hereafter, client-server communication is used to update the local models.
\begin{algorithm}[H]
    \caption{LTH training server-side}\label{alg:server Federated}
    \begin{algorithmic}[1]
        \Require k, $flag = 1$, prune\_rate; \textit{Client\hspace{2px}i\hspace{1px}}: $\theta_{r}^{i}$
        \Ensure Global Parameters: ${\Theta_{}^{*}}$ 
        
        \hspace{-3em}//Server selects k-client parameters from $clients\_pool$. 
        \State \textit{\textbf{Aggregation:}} 
        
         \hspace{3em}    ${\Theta_{r+1} \leftarrow \frac{1}{k} \sum_{i}^{i=k} \theta_{r}^{i} }$
 
        \raggedright
        \State \textit{\textbf{Sparsification:}} \newline
            \hspace{1em}  {\textbf{if} $clients\_pool ==$ "High" {\textbf{then}} }\\ \hspace{1.5em} {\textit{No sparsification}}\newline
            \hspace{1em}  {\textbf{if} $clients\_pool ==$ "Mid" and $flag == 1$ {\textbf{then}} }\\ \hspace{1.5em} {\textit{Start sparsifying} with $prune\_rate = p/100$}\\
            \hspace{1.5em} \textit{Broadcast prune\_mask to Clients} \\
            \hspace{1.5em} \textit{Increment $flag$ by $1$}\newline
            \hspace{1em}  {\textbf{if} $clients\_pool ==$ "Low" and $flag == 2$ {\textbf{then}} }\\ \hspace{1.5em} {\textit{Start sparsifying} with $prune\_rate = p_{new}/100$}\\
            \hspace{1.5em} \textit{Broadcast prune\_mask to Clients} \\
            \hspace{1.5em} \textit{Increment $flag$ by $1$}\\  
        \State \textbf{Broadcast to Clients:} ${\Theta_{r+1}}$ \\
    \end{algorithmic}
\end{algorithm}

Sparsifying of the parameters acts by changing $1$ to $0$ in a binary mask that disables the gradient flow. In the binary mask, $1$ indicates the active weights, while $0$ indicates inactive weights. The prune\_rate hyperparameter determines the number of parameters to be sparsed. The prune\_mask with dropped weights is computed as illustrated in Algorithm \ref{alg:prune}.
\begin{algorithm}[H]
    \caption{sparsification}\label{alg:prune}
    \begin{algorithmic}[1]
        \Require { \textit{weight, prune\_rate}}
        \Ensure {prune\_mask}
        \State \textit{ \# Count zeroes in aggregated weights at the server.} \newline
             ${num\_zeros \leftarrow (weight==0).sum()}$ \newline
             ${num\_nonzeros \leftarrow weight.count() - num\_zeros}$
        
        \State \textit{ \# Number of parameters to be sparsed.} \newline
            ${num\_prune \leftarrow prune\_rate * num\_nonzeros}$ \newline
       
        \If {num\_prune == 0:} 
            \State {\textbf{return} weight.bool()}
        \EndIf
        \State {$p \leftarrow num\_zeros + num\_prune$}
        \State {$w, id_w \leftarrow sort(weight)$}
        \State {$prune\_mask \leftarrow weight.bool().int()$}
        \State {$prune\_mask.data.view(-1)[id_w[:p]] \leftarrow 0 $}
        \State {\textbf{return} prune\_mask} \\
        \End
    \end{algorithmic}
\end{algorithm}

\section{Results and Discussion}
\subsection{Dataset Description}
We demonstrate the performance of Equitable-FL on three datasets: $MNIST$, $F-MNIST$, and $CIFAR-10$. The $MNIST$ dataset is a well-known collection of handwritten digits, while the $F-MNIST$ dataset contains fashion and lifestyle images. Both datasets, which are used to train various image recognition models, have the same image size and train-test split. Specifically, these datasets include a training set of $60,000$ samples and a test set of $10,000$ samples. Each sample is a $28\times28$ grayscale image associated with a label from one of the ten distinct classes. In comparison, the $CIFAR-10$ dataset is a subset of the Tiny Images dataset and contains images from $10$ classes, comprising $500$ training images and $100$ testing images for every class; each sample is an RGB $32\times32$ pixels image. 

$Brain-MRI$ dataset contains $7023$ images of human brain MRI to classify the brain into four categories: glioma, meningioma, no tumour and pituitary.  A brain tumour is the growth of abnormal cells in the brain. It can be cancerous (malignant) or noncancerous (benign). The images in this dataset are different, so we resized the image to an RGB $120\times120$ pixels. Medical information is extremely sensitive and is governed by strict privacy regulations. Also, medical data is dynamic, and more data becomes available as time passes. Without data centralization, federated learning enables machine learning models to be trained directly on data sources (such as hospitals and clinics) and promotes continuous learning. As new knowledge and data evolve, models can change and advance over the iterative process.

$PlantVillage$ dataset contains $20,600$ leaf images of healthy and diseased plants that are categorized into $15$ different classes, with the size of each image to an RGB $256\times256$ pixels. Data on plants is typically derived from several demographic circumstances. This variety may improve the machine learning models' robustness and generalizability. Data from several sources can be trained through federated learning to acquire more comprehensive learning.

\subsection{Model Setup }
For our experiments, the Client uses CNN to train and shares the updates to the server for aggregation. We form two separate scenarios that differ in architectural configurations; dense and sparse. The convolutional, max-pooling, and fully connected layers are dense in the first configuration. However, in the second configuration, networks are sparsified by dropping off some of the convolutional and fully connected connections. 
Generally, two main events are performed by the server, one is for selecting $k$-clients, and the other is aggregation for global parameter calculation. Here, the server has the additional responsibility of sparsifying the network parameters. 


\subsection{Model training}
Model training takes place over multiple phases in which nodes with similar heterogeneity score participate in one phase. In the initial phase, the model has all parameters and nodes that have high heterogeneity score that have the capability to train the model and communicate the parameters that participate in the process. At the end of the first phase, the trained model is pruned. Nodes with lesser heterogeneity score join the training. This process of progressive model pruning, onboarding of new nodes and training continues in subsequent phases depending on the number of nodes and variation in their heterogeneity score. In the present work, training is done in three phases. The nodes are categorized in three groups based on heterogeneity score computed by primary memory for computation, secondary memory for storage, floating-point unit to carry out operations on floating point numbers and bandwidth. Nodes with high heterogeneity score participate in training in the first phase, nodes with average score join in the second phase and other nodes are allowed in the third phase of training.

\subsection{Results}
We experimentally provide a solution for federated learning in a resource-constrained environment. Our goal is not to achieve state-of-the-art accuracy but to solve the existing heterogeneity problem. We focus on situations where clients have high, moderate and lower heterogeneity scores.\\ 

\begin{figure}[H]
    \centering
    \begin{subfigure}{.33\textwidth}
        \centering
        \includegraphics[width=1.1\textwidth]{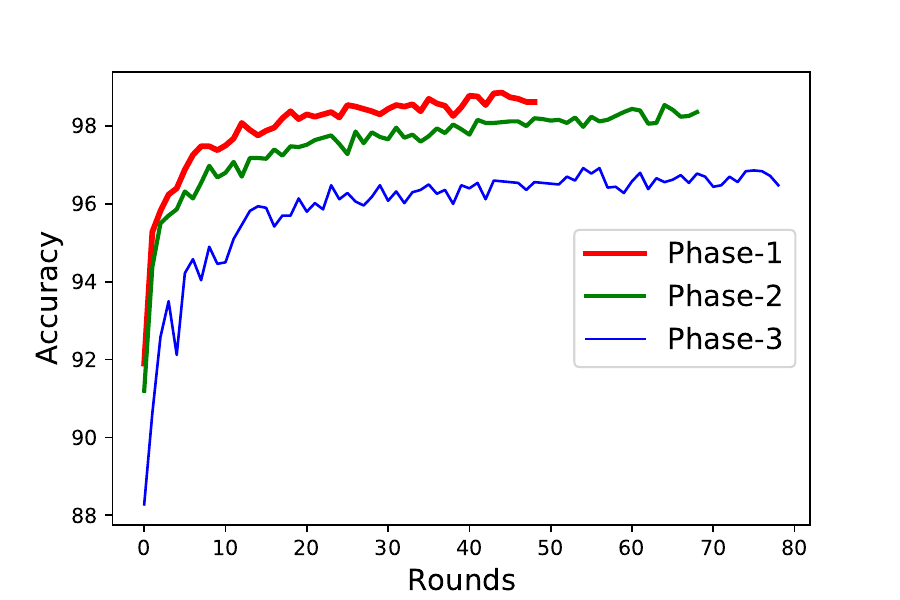}
        \caption[short]{$MNIST$}
    \end{subfigure}%
    \begin{subfigure}{.33\textwidth}
        \centering
        \includegraphics[width=1.1\textwidth]{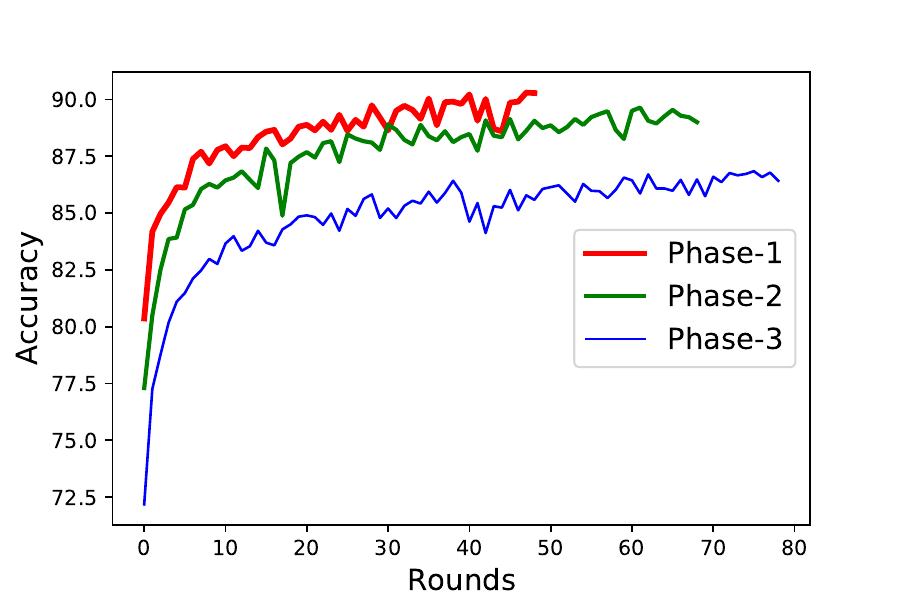}
        \caption[short]{$F-MNIST$}
    \end{subfigure}%
    \hfill
    \begin{subfigure}{.33\textwidth}
        \centering
        \includegraphics[width=1.1\textwidth]{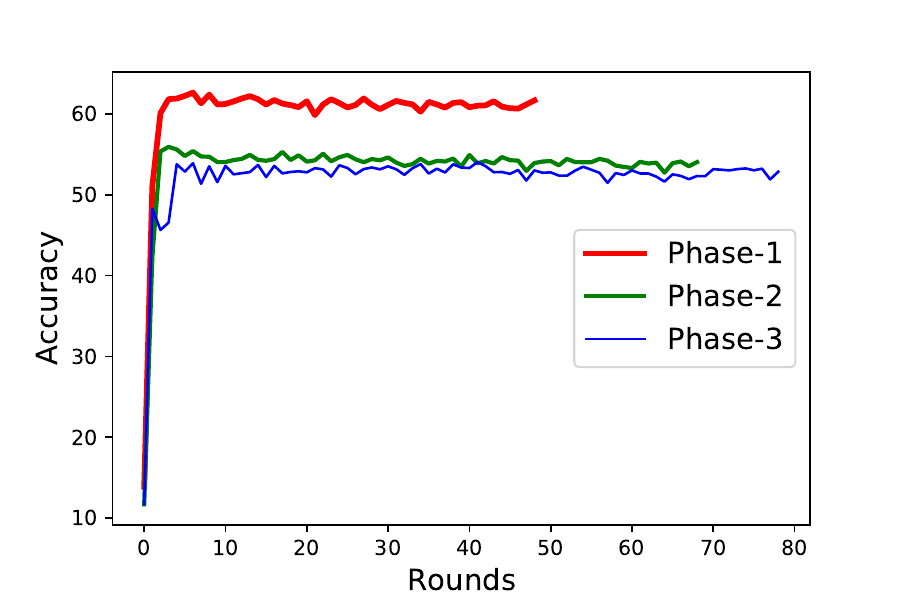}
        \caption[short]{$CIFAR-10$}
    \end{subfigure}%
    \begin{subfigure}{.33\textwidth}
        \centering
        \includegraphics[width=1.1\textwidth]{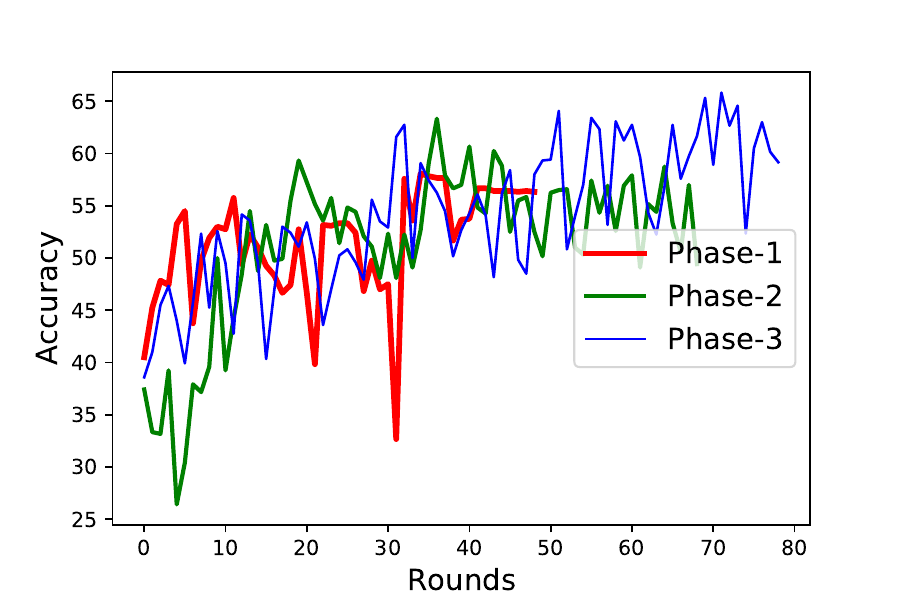}
        \caption[short]{$Brain-MRI$}
    \end{subfigure}%
    \hfill
    \begin{subfigure}{.33\textwidth}
        \centering
        \includegraphics[width=1.1\textwidth]{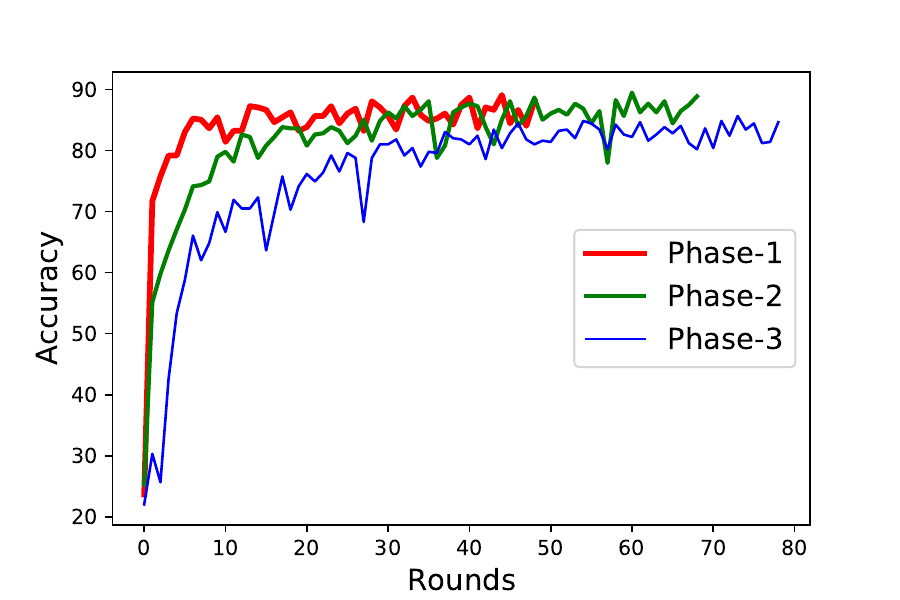}
        \caption[short]{$PlantVillage$}
    \end{subfigure}%
\caption[short]{\textbf{Equitable-FL performance during training per round for three phases.}} \label{fig:f3}
\end{figure}
\vspace*{-1.5em}
The improvement in testing accuracy over the communication round for all datasets during the training in three different phases are shown in Figure \ref{fig:f3}. The performance of lightweight FL with the LTH  is comparable to vanilla FL. We additionally observe that FL with the LTH needs a few more rounds for proximate accuracy. The total number of communication rounds used to evaluate models is $50$, $70$, and $80$ during the first, second and third phases, respectively. The second and third phases show a marginal drop in accuracy compared to the first phase for the $MNIST$ and $F-MNIST$ datasets, where sparsity increased to $80\%$ and $70\%$, respectively. Conversely, performance on the $CIFAR-10$ dataset decreases significantly over the increase in sparsity using The LTH approach. However, $Brain-MRI$ and $PlantVillage$ dataset performs nicely over the same approach, with sparsity increased to $50\%$ and $80\%$, respectively. \\

\textbf{Performance of the LTH on FL:} We evaluate the LTH for FL, using the distinct model architectures, for three benchmark datasets, $MNIST$, $F-MNIST$ and $CIFAR-10$, and two application-specific datasets, $Brain-MRI$ data and $PlantVillage$ data. In Table \ref{table:t1}, we show the performance and the sparsity percentage of models with their settings in all three phases: 
(1) In the first phase, the performance of vanilla FL on all mentioned datasets are $98.8\%$, $90.3\%$, $62.7\%$, $58.1\%$, and $89.3\%$, respectively. (2) In the next phase, when we employ the LTH approach for the first time, the performance of $30\%$ sparse FL on the $MNIST$, $F-MNIST$, $Brain-MRI$ and $PlantVillage$ datasets is $98.5\%$, $89.6\%$, $63.3\%$, $90.9\%$ respectively, and the performance of 20\% sparse FL on the $CIFAR-10$ dataset is $56.0\%$. (3) In the last phase, we employ the LTH approach again for additional sparsity of models, resulting in $80\%$ sparse FL on the $MNIST$ and $PlantVillage$ datasets with the accuracy of $96.9\%$ and $89.5\%$ respectively, for $50\%$, $60\%$, and $70\%$ sparse FL on the $Brain-MRI$, $CIFAR-10$ and $F-MNIST$ datasets exhibit $65.8\%$, $54.1\%$, and $86.8\%$ accuracy respectively. 

\begin{table}[!hbt]
\begin{subtable}[t]{\textwidth}
\centering
\begin{tabular}[t]{|c|c|c||c|c|c|c|c|}
\hline
\multicolumn{3}{|c||}{~} & \multicolumn{5}{c|}{Phase 1}\\ \hline
S. No. & Dataset & \#Samples & Rounds & \# Nodes & LTH & Sparsity(\%) & Acc.(\%) \\ \hline
1 & {$MNIST$} & 500 & \multirow{5}{*}{50} & 30 & \multirow{5}{*}{No} & \multirow{5}{*}{0} & 98.8  \\ \cline{1-3}\cline{5-5}
2 & {$F-MNIST$} & 500 & ~ & 30 & ~ & ~ & 90.3 \\ \cline{1-3}\cline{5-5}
3 & {$CIFAR-10$} & 5000 & ~ & 6 & ~ & ~ & 62.7 \\ \cline{1-3}\cline{5-5}
4 & {$Brain-MRI$} & 320 & ~ & 9 & ~ & ~ & 58.1   \\ \cline{1-3}\cline{5-5}
5 & {$PlantVillage$} & 500 & ~ & 6 & ~ & ~ & 89.3 \\ \hline
\end{tabular}
\caption{\footnotesize Performance of the model without sparsity (Vanilla FL) in the first phase.}
\label{tab:table1_a}
\end{subtable}

\begin{subtable}[t]{\textwidth}
\centering
\begin{tabular}[t]{|c|c|c||c|c|@{}c@{}|c|c|}
\hline
\multicolumn{3}{|c||}{~} & \multicolumn{5}{c|}{Phase 2}\\ \hline
S. No. & Dataset & \#Samples & Rounds & \# Nodes & LTH & Sparsity(\%) & Acc.(\%) \\ \hline
1 & {$MNIST$} & 500 & \multirow{5}{*}{70} & 60 & \multirow{5}{*}{$1^{st}$ time} & 30 & 98.5  \\ \cline{1-3}\cline{5-5}
2 & {$F-MNIST$} & 500 & ~ & 60 & ~ & 30 & 89.6 \\ \cline{1-3}\cline{5-5}
3 & {$CIFAR-10$} & 5000 & ~ & 12 & ~ & 20 & 56.0 \\ \cline{1-3}\cline{5-5}
4 & {$Brain-MRI$} & 320 & ~ & 18 & ~ & 30 & 63.3   \\ \cline{1-3}\cline{5-5}
5 & {$PlantVillage$} & 500 & ~ & 12 & ~ & 30 & 90.9 \\ \hline
\end{tabular}
\caption{\footnotesize Performance of the model with moderate sparsity in the second phase.}
\label{tab:table1_b}
\end{subtable}

\begin{subtable}[t]{\textwidth}
\centering
\begin{tabular}[t]{|c|c|c||c|c|@{}c@{}|c|c|}
\hline
\multicolumn{3}{|c||}{~} & \multicolumn{5}{c|}{Phase 3}\\ \hline
S. No. & Dataset & \#Samples & Rounds & \# Nodes & LTH & Sparsity(\%) & Acc.(\%) \\ \hline
1 & {$MNIST$} & 500 & \multirow{5}{*}{80} & 100 & \multirow{5}{*}{$2^{nd}$ time} & 80 & 96.9  \\ \cline{1-3}\cline{5-5}
2 & {$F-MNIST$} & 500 & ~ & 100 & ~ & 70 & 86.8 \\ \cline{1-3}\cline{5-5}
3 & {$CIFAR-10$} & 5000 & ~ & 20 & ~ & 60 & 54.1 \\ \cline{1-3}\cline{5-5}
4 & {$Brain-MRI$} & 320 & ~ & 30 & ~ & 50 & 65.8   \\ \cline{1-3}\cline{5-5}
5 & {$PlantVillage$} & 500 & ~ & 20 & ~ & 80 & 89.5 \\ \hline

\end{tabular}
\caption{\footnotesize Performance of the model for an increase in sparsity in the third phase.}
\label{tab:table1_c}
\end{subtable}

\caption{\footnotesize \textbf{The performance of the Equitable-FL in terms of accuracy during phase 1 \subref{tab:table1_a}, phase 2 \subref{tab:table1_b} and phase 3 \subref{tab:table1_c} on $MNIST$, $F-MNIST$, $CIFAR-10$, $Brain-MRI$ and $PlantVillage$ datasets.}}
\label{table:t1}
\end{table}

\textbf{Effect of the LTH on FL for model size compaction:} We figured model size compaction after applying the LTH on FL, and it gives the advantage of space-saving while communicating weights during FL training. Table \ref{table:t2} shows the observed effects on the compaction of model size. We achieved space savings of  $21.7\%$, $21.7\%$, $15.2\%$, $24.5\%$ and $24.4\%$ for $MNIST$, $F-MNIST$ and $CIFAR-10$, $Brain-MRI$ data and $PlantVillage$ data, respectively, during training in the second phase. Subsequently, it gained $65.2\%$, $56.5\%$, $52.5\%$, $55.4\%$ and $71.4\%$ of space saving in the third phase for respective datasets. 

\begin{table}[H]
\begin{subtable}[t]{\textwidth}
\centering
\begin{tabular}[t]{|c|c||c|c|}
\hline
\multicolumn{2}{|c||}{~} & \multicolumn{2}{c|}{Phase 1}\\ \hline
\multicolumn{1}{|p{0.5cm}|}{\centering  \vspace{0.1em} S. No. } & \multicolumn{1}{p{1.5cm}||}{\centering  \vspace{0.1em}Dataset } & \multicolumn{1}{p{1.5cm}|}{\centering \vspace{0.1em} \char"0023  NNZ \\Parameters} & \multicolumn{1}{p{1.6cm}|}{\centering Compressed \\ Model \\ Size (KB) } \\ \hline
1 & $MNIST$ & 5,882 & 23 \\ \hline
2 & $F-MNIST$ & 5,882 & 23 \\ \hline
3 & $CIFAR-10$ & 1,63,02,086 & 58,560 \\ \hline
4 & $Brain-MRI$ & 20,14,044 & 7,247 \\ \hline 
5 & $PlantVillage$ & 1,34,367 & 487 \\ \hline
\end{tabular}
\caption{\footnotesize Description of Non-zero parameters and compressed size of the model in Vanilla FL.}
\label{tab:table2_a}
\end{subtable}

\begin{subtable}[t]{\textwidth}
\centering
\begin{tabular}[t]{|c|c||c|c|c|c|}
\hline
\multicolumn{2}{|c||}{~} & \multicolumn{4}{c|}{Phase 2} \\ \hline
\multicolumn{1}{|p{0.5cm}|}{\centering  \vspace{0.1em} S. No. } & \multicolumn{1}{p{1.5cm}||}{\centering  \vspace{0.1em}Dataset } & \multicolumn{1}{p{1.5cm}|}{\centering \vspace{0.1em} \# NNZ \\ Parameters} & \multicolumn{1}{p{1.5cm}|}{\centering \vspace{0.1em} \# Zero \\ Parameters} & \multicolumn{1}{p{1.6cm}|}{\centering Compressed \\ Model \\ Size (KB) } & \multicolumn{1}{p{1.55cm}|}{\centering  \vspace{0.1em} Space\\ Saving(\%)} \\ \hline
1 & $MNIST$ & 4,113 & 1,769 & 18 & 21.7  \\ \hline
2 & $F-MNIST$ & 4,113 & 1,769 & 18 & 21.7 \\ \hline
3 & $CIFAR-10$ & 1,30,41,832 & 32,60,254 & 49,631 & 15.2 \\ \hline
4 & $Brain-MRI$ & 14,10,020 & 6,04,024 & 5,468 & 24.5 \\ \hline
5 & $PlantVillage$ & 94,054 & 40,313 & 368 & 24.4 \\ \hline

\end{tabular}
\caption{\footnotesize Description of Non-zero parameters, compressed size and space-saving of the model in the second phase.}
\label{tab:table2_b}
\end{subtable}

\begin{subtable}[t]{\textwidth}
\centering
\begin{tabular}[t]{|c|c||c|c|c|c|}
\hline
\multicolumn{2}{|c||}{~} & \multicolumn{4}{c|}{Phase 3} \\ \hline
\multicolumn{1}{|p{0.5cm}|}{\centering  \vspace{0.1em} S. No. } & \multicolumn{1}{p{1.5cm}||}{\centering  \vspace{0.1em}Dataset } & \multicolumn{1}{p{1.5cm}|}{\centering \vspace{0.1em} \# NNZ \\ Parameters} & \multicolumn{1}{p{1.5cm}|}{\centering \vspace{0.1em} \# Zero \\ Parameters} & \multicolumn{1}{p{1.6cm}|}{\centering Compressed \\ Model \\ Size (KB) } & \multicolumn{1}{p{1.55cm}|}{\centering  \vspace{0.1em} Space\\ Saving(\%)} \\ \hline
1 & $MNIST$ & 1,261 & 4,621 & 8 & 65.2  \\ \hline
2 & $F-MNIST$ & 1,832 & 4,050 & 10 & 56.5 \\ \hline
3 & $CIFAR-10$ & 63,70,002 & 99,32,084 & 27,778 & 52.5 \\ \hline
4 & $Brain-MRI$ & 10,07,339 & 10,06,705 & 4021 & 55.4\\ \hline
5 & $PlantVillage$ & 26,962 & 1,07,405 & 139 & 71.4\\ \hline
\end{tabular}
\caption{\footnotesize Description of Non-zero parameters, compressed size and space saving of the model in the third phase.}
\label{tab:table2_c}
\end{subtable}

\caption{\footnotesize \textbf{Overview of Non-zero parameters, compressed size and space-saving of the Equitable-FL during phase 1 \subref{tab:table2_a}, phase 2 \subref{tab:table2_b} and phase 3 \subref{tab:table2_c} on $MNIST$, $F-MNIST$, $CIFAR-10$, $Brain-MRI$ and $PlantVillage$ datasets.}}
\label{table:t2}
\end{table}

\begin{table}[H]
    \centering
    \begin{tabular}{|c|c||c|c||c|c|}
    \hline
        \multicolumn{2}{|c||}{~} & \multicolumn{2}{c||}{\textbf{Phase 2}} & \multicolumn{2}{c|}{\textbf{Phase 3}} \\ \hline
        \textbf{S. No.} & \textbf{Dataset} & \textbf{Speed-up} & \textbf{Acc.(\%)} & \textbf{Speed-up} & \textbf{Acc.(\%)} \\ \hline
        1 & $MNIST$ & 1.43x & 98.5 & 4.66x & 96.9 \\ \hline
        2 & $F-MNIST$ & 1.43x & 89.6 & 3.21x & 86.8 \\ \hline
        3 & $CIFAR-10$ & 1.25x & 56.0 & 2.56x & 54.1 \\ \hline
        4 & $Brain-MRI$ & 1.43x & 63.3 & 1.99x & 65.8 \\ \hline
        5 & $PlantVillage$ & 1.43x & 90.9 & 4.98x & 89.5 \\ \hline
    \end{tabular}
    \caption{\normalsize \textbf{Speed-up gained during the sparsification in the first and second phase on all five datasets.}} \label{table:t3}
\end{table}

\textbf{Effect of the LTH on FL for lowering operations on training:} After applying the LTH to FL, we computed execution speed-up using equation \ref{eq:Speed-up} for training different dataset-specific models. Table \ref{table:t3} depicts the computed speed-up value. We gained speed-up of  $1.43$x, $1.43$x, $1.25$x, $1.43$x, and $1.43$x in the second phase and $4.66$x, $3.21$x, $2.56$x, $1.99$x, and  $4.98$x in the third phase for $MNIST$, $F-MNIST$ and $CIFAR-10$, $Brain-MRI$ data and $PlantVillage$ data, respectively.
\subsection{Discussion}
The findings support the objective of Equitable-FL, which is to gradually use the LTH in the FL to promote weaker client participation in a resource-constrained environment. Figure \ref{fig:f3} shows the training performed during three phases. It demonstrates that the model needs more rounds to achieve comparable accuracy over the increase in sparsity. The effectiveness of the Equitable-FL can be examined on two factors:
\begin{itemize}
    \item Space-saving
    \item Speed-up
\end{itemize}
The FL is affected by resource heterogeneity and scarcity, which is quantified by the heterogeneity score. Using this, model training over multiple phases yields remarkable performance. Three benchmark datasets, $MNIST$, $F-MNIST$, and $CIFAR-10$, as well as application-specific datasets for $PlantVillage$ and $Brain-MRI$ imaging, are used to validate the Equitable-FL. The findings illustrate that accuracy remains very close, but at the same time, space-saving is gained. $Brain-MRI$ and $PlantVillage$ datasets give better accuracy over the sparsifying to achieve higher space-saving and speed-up. Unlike the problem statement, when there is no problem of resource heterogeneity or scarcity, the model reduces to vanilla FL with only one phase. We observed that the Equitable-FL is able to match the model size with the resource requirement to overcome the heterogeneity and scarcity problem.

\section{Conclusion}

This work exploited the LTH progressively in the FL to facilitate weaker client participation in a resource-constrained environment. An extensive analysis of the Equitable-FL was presented for application-specific datasets, $Brain-MRI$ data and $PlantVillage$ data, along with three benchmark datasets $MNIST$, $F-MNIST$ and $CIFAR-10$. The performance was evaluated in a phased manner to draw the effectiveness of the Equitable-FL. The results showed that accuracy remains close but found phenomenal model space-saving, except for the $CIFAR-10$ dataset. A significant loss in accuracy was observed for model space saving of $15.2\%$ and $52.5\%$ in the second and third phases, respectively, in $CIFAR-10$. $Brain-MRI$ and $PlantVillage$ datasets give an accuracy of $58.1\%$ and $89.3\%$ in the first phase, $63.3\%$ and $90.9\%$ on space-saving of $24.5\%$ and $24.4\%$ in the second phase, and $65.8\%$ and $89.5\%$ on space-saving of $25.4\%$ and $71.4\%$ in the third phase. Sparsity reduces the floating points for computation during training and achieves a speed-up of $1.43$x for both datasets in the second phase and $1.99$x and $4.98$x in the third phase, respectively. The results confirmed the eﬀectiveness of the Equitable-FL in a resource-constrained environment that is characterized by resource heterogeneity and scarcity. Equitable-FL was able to match the accuracy of vanilla FL with a reduced model size and increased speed-up. However, some resource rich nodes may have large datasets that can cause long delays in training periods. Equitable-FL needs to be modified to cater to such disparity in size of data sets on nodes. 

\bibliographystyle{unsrt}  
\bibliography{references}  

\begin{thebibliography}{10}

\bibitem{deep_survey_3}
Absalom~E Ezugwu, Abiodun~M Ikotun, Olaide~O Oyelade, Laith Abualigah,
  Jeffery~O Agushaka, Christopher~I Eke, and Andronicus~A Akinyelu.
\newblock A comprehensive survey of clustering algorithms: State-of-the-art
  machine learning applications, taxonomy, challenges, and future research
  prospects.
\newblock {\em Engineering Applications of Artificial Intelligence},
  110:104743, 2022.

\bibitem{unsupervised_learning_1}
Amir~H Gandomi, Fang Chen, and Laith Abualigah.
\newblock Machine learning technologies for big data analytics, 2022.

\bibitem{deep_learning_1}
Shi Dong, Ping Wang, and Khushnood Abbas.
\newblock A survey on deep learning and its applications.
\newblock {\em Computer Science Review}, 40:100379, 2021.

\bibitem{federated_healthcare}
Jie Xu, Benjamin~S Glicksberg, Chang Su, Peter Walker, Jiang Bian, and Fei
  Wang.
\newblock Federated learning for healthcare informatics.
\newblock {\em Journal of Healthcare Informatics Research}, 5(1):1--19, 2021.

\bibitem{federated_healthcare_2}
Alexander Chowdhury, Hasan Kassem, Nicolas Padoy, Renato Umeton, and Alexandros
  Karargyris.
\newblock A review of medical federated learning: Applications in oncology and
  cancer research.
\newblock In {\em Brainlesion: Glioma, Multiple Sclerosis, Stroke and Traumatic
  Brain Injuries: 7th International Workshop, BrainLes 2021, Held in
  Conjunction with MICCAI 2021, Virtual Event, September 27, 2021, Revised
  Selected Papers, Part I}, pages 3--24. Springer, 2022.

\bibitem{federated_learning_survey_2}
Syreen Banabilah, Moayad Aloqaily, Eitaa Alsayed, Nida Malik, and Yaser
  Jararweh.
\newblock Federated learning review: Fundamentals, enabling technologies, and
  future applications.
\newblock {\em Information Processing \& Management}, 59(6):103061, 2022.

\bibitem{federated_learning_survey_1}
Fahad~Ahmed KhoKhar, Jamal~Hussain Shah, Muhammad~Attique Khan, Muhammad
  Sharif, Usman Tariq, and Seifedine Kadry.
\newblock A review on federated learning towards image processing.
\newblock {\em Computers and Electrical Engineering}, 99:107818, 2022.

\bibitem{federated_survey4}
Ruchi Gupta and Tanweer Alam.
\newblock Survey on federated-learning approaches in distributed environment.
\newblock {\em Wireless Personal Communications}, 125(2):1631--1652, 2022.

\bibitem{federated_taxonomy}
Qiang Yang, Yang Liu, Tianjian Chen, and Yongxin Tong.
\newblock Federated machine learning: Concept and applications.
\newblock {\em ACM Trans. Intell. Syst. Technol.}, 10(2), jan 2019.

\bibitem{DeepL}
Yann LeCun, Yoshua Bengio, and Geoffrey Hinton.
\newblock Deep learning.
\newblock {\em nature}, 521(7553):436--444, 2015.

\bibitem{sparse_deep_1}
Rongrong Ma and Lingfeng Niu.
\newblock A survey of sparse-learning methods for deep neural networks.
\newblock In {\em 2018 IEEE/WIC/ACM International Conference on Web
  Intelligence (WI)}, pages 647--650. IEEE, 2018.

\bibitem{sparse_deep_2}
Yingjie Tian and Yuqi Zhang.
\newblock A comprehensive survey on regularization strategies in machine
  learning.
\newblock {\em Information Fusion}, 80:146--166, 2022.

\bibitem{sparse_deep_3}
Licheng Jiao, Yuting Yang, Fang Liu, Shuyuan Yang, and Biao Hou.
\newblock The new generation brain-inspired sparse learning: A comprehensive
  survey.
\newblock {\em IEEE Transactions on Artificial Intelligence}, 2022.

\bibitem{Sparsity-in-Deep-Learning}
Torsten Hoefler, Dan Alistarh, Tal Ben{-}Nun, Nikoli Dryden, and Alexandra
  Peste.
\newblock Sparsity in deep learning: Pruning and growth for efficient inference
  and training in neural networks.
\newblock {\em CoRR}, abs/2102.00554, 2021.

\bibitem{Sparse-Networks-from-Scratch}
Tim Dettmers and Luke Zettlemoyer.
\newblock Sparse networks from scratch: Faster training without losing
  performance.
\newblock {\em CoRR}, abs/1907.04840, 2019.

\bibitem{federated_sparse1}
Sameer Bibikar, Haris Vikalo, Zhangyang Wang, and Xiaohan Chen.
\newblock Federated dynamic sparse training: Computing less, communicating
  less, yet learning better.
\newblock In {\em Proceedings of the AAAI Conference on Artificial
  Intelligence}, volume~36, pages 6080--6088, 2022.

\bibitem{malach2020proving}
Eran Malach, Gilad Yehudai, Shai Shalev-Schwartz, and Ohad Shamir.
\newblock Proving the lottery ticket hypothesis: Pruning is all you need.
\newblock In {\em International Conference on Machine Learning}, pages
  6682--6691. PMLR, 2020.

\bibitem{BMRI}
Msoud Nickparvar.
\newblock Brain tumor mri dataset, 2021.

\bibitem{PlantData}
Tairu~Oluwafemi Emmanuel.
\newblock Plantvillage dataset, 2018.

\bibitem{pmlr-v54-mcmahan17a}
Brendan McMahan, Eider Moore, Daniel Ramage, Seth Hampson, and Blaise Aguera~y
  Arcas.
\newblock {Communication-Efficient Learning of Deep Networks from Decentralized
  Data}.
\newblock In Aarti Singh and Jerry Zhu, editors, {\em Proceedings of the 20th
  International Conference on Artificial Intelligence and Statistics},
  volume~54 of {\em Proceedings of Machine Learning Research}, pages
  1273--1282. PMLR, 20--22 Apr 2017.

\bibitem{LTH2018}
Jonathan Frankle and Michael Carbin.
\newblock The lottery ticket hypothesis: Training pruned neural networks.
\newblock {\em CoRR}, abs/1803.03635, 2018.

\bibitem{DBLP:journals/corr/abs-1911-11134}
Utku Evci, Trevor Gale, Jacob Menick, Pablo~Samuel Castro, and Erich Elsen.
\newblock Rigging the lottery: Making all tickets winners.
\newblock {\em CoRR}, abs/1911.11134, 2019.

\bibitem{pmlr-v119-tan20a}
Chong Min~John Tan and Mehul Motani.
\newblock {D}rop{N}et: Reducing neural network complexity via iterative
  pruning.
\newblock In Hal~Daumé III and Aarti Singh, editors, {\em Proceedings of the
  37th International Conference on Machine Learning}, volume 119 of {\em
  Proceedings of Machine Learning Research}, pages 9356--9366. PMLR, 13--18 Jul
  2020.

\bibitem{10.1145/3307650.3322263}
Jiaqi Zhang, Xiangru Chen, Mingcong Song, and Tao Li.
\newblock Eager pruning: Algorithm and architecture support for fast training
  of deep neural networks.
\newblock In {\em Proceedings of the 46th International Symposium on Computer
  Architecture}, ISCA '19, page 292–303, New York, NY, USA, 2019. Association
  for Computing Machinery.

\bibitem{10.1145/3369583.3392681}
Linnan Wang, Wei Wu, Junyu Zhang, Hang Liu, George Bosilca, Maurice Herlihy,
  and Rodrigo Fonseca.
\newblock Fft-based gradient sparsification for the distributed training of
  deep neural networks.
\newblock In {\em Proceedings of the 29th International Symposium on High
  Performance Parallel and Distributed Computing}, HPDC '20, page 113–124,
  New York, NY, USA, 2020. Association for Computing Machinery.

\bibitem{DBLP:journals/corr/abs-1905-01067}
Hattie Zhou, Janice Lan, Rosanne Liu, and Jason Yosinski.
\newblock Deconstructing lottery tickets: Zeros, signs, and the supermask.
\newblock {\em CoRR}, abs/1905.01067, 2019.

\bibitem{DBLP:journals/corr/abs-1711-01263}
Jingyang Zhu, Jingbo Jiang, Xizi Chen, and Chi{-}Ying Tsui.
\newblock Sparsenn: An energy-efficient neural network accelerator exploiting
  input and output sparsity.
\newblock {\em CoRR}, abs/1711.01263, 2017.

\bibitem{https://doi.org/10.48550/arxiv.1710.01878}
Michael Zhu and Suyog Gupta.
\newblock To prune, or not to prune: exploring the efficacy of pruning for
  model compression, 2017.

\bibitem{federated_nlp_1}
Ming Liu, Stella Ho, Mengqi Wang, Longxiang Gao, Yuan Jin, and He~Zhang.
\newblock Federated learning meets natural language processing: A survey.
\newblock {\em arXiv preprint arXiv:2107.12603}, 2021.

\bibitem{personalized_federated}
Alysa~Ziying Tan, Han Yu, Lizhen Cui, and Qiang Yang.
\newblock Towards personalized federated learning.
\newblock {\em IEEE Transactions on Neural Networks and Learning Systems},
  2022.

\bibitem{federated_sparse2}
Shan Ullah and Deok-Hwan Kim.
\newblock Federated learning using sparse-adaptive model selection for embedded
  edge computing.
\newblock {\em IEEE Access}, 9:167868--167879, 2021.

\bibitem{federated_sparse3}
Qianqian Tong, Guannan Liang, Tan Zhu, and Jinbo Bi.
\newblock Federated nonconvex sparse learning.
\newblock {\em arXiv preprint arXiv:2101.00052}, 2020.

\end{thebibliography}
\nocite{*}

\end{document}